\documentclass[11pt,a4paper]{article}
\usepackage{authblk}
\usepackage[hyperref]{emnlp-ijcnlp-2019}
\usepackage{times}
\usepackage{latexsym}
\usepackage{adjustbox}
\usepackage{multirow}
\usepackage{array}
\usepackage{arydshln}
\usepackage{amsthm}

\usepackage{url}
\usepackage{graphicx}
\usepackage{caption}
\usepackage{subcaption}
\usepackage{array}
\usepackage{scrextend}
\usepackage{footnote}
\makesavenoteenv{tabular}
\makesavenoteenv{table}
\usepackage{breqn}

\aclfinalcopy 

\def\JNdel#1{\bgroup\markoverwith{\textcolor{green}{\rule[0.5ex]{2pt}{1pt}}}\ULon{#1}}

\def\KSdel#1{\bgroup\markoverwith{\textcolor{red}{\rule[0.5ex]{2pt}{1pt}}}\ULon{#1}}

\def\ABdel#1{\bgroup\markoverwith{\textcolor{blue}{\rule[0.5ex]{2pt}{1pt}}}\ULon{#1}}

\title{Robustness of Linguistic Features:\\Syntactic Errors are More Important than Lexical}
\title{Lexical features are more vulnerable,\\ syntactic features have more predictive power}
\title{Lexical Features Are More Vulnerable,\\Syntactic Features Have More Predictive Power}

\author[1]{\textbf{Jekaterina Novikova}}
\author[1]{\textbf{Aparna Balagopalan}}
\author[2]{\textbf{Ksenia Shkaruta}}
\author[1,3]{\textbf{Frank Rudzicz}}
\affil[1]{Winterlight Labs, \{jekaterina, aparna\}@winterlightlabs.com}
\affil[2]{Georgia Tech, ksenia.shkaruta@gatech.edu}
\affil[3]{University of Toronto; Vector Institute for Artificial Intelligence, frank@cs.toronto.edu}

\date{}

\begin{document}
\maketitle
\begin{abstract}
Understanding the vulnerability of linguistic features extracted from noisy text is important for both developing better health text classification models and for interpreting vulnerabilities of natural language models. In this paper, we investigate how generic language characteristics, such as syntax or the lexicon, are impacted by artificial text alterations. The vulnerability of features is analysed from two perspectives: (1) the level of feature value change, and (2) the level of change of feature predictive power as a result of text modifications. We show that lexical features are more sensitive to text modifications than syntactic ones. However, we also demonstrate that these smaller changes of syntactic features have a stronger influence on classification performance downstream, compared to the impact of changes to lexical features. Results are validated across three datasets representing different text-classification tasks, with different levels of lexical and syntactic complexity of both conversational and written language.
\end{abstract}

\section{Introduction}

It is important to understand the vulnerability of linguistic features to text alteration because (1) pre-defined linguistic features are still frequently used in health text classification, e.g., detecting Alzheimer’s disease (AD)  \cite{ad_masrani-etal-2017-detecting, zhu2018deconfounding, balagopalan2018effect}, aphasia \cite{aph_fraser2015sentence}, or sentiment from language~\cite{sent_maas2011learning}; and (2) understanding the importance of syntactic and lexical information separately as well as interactively is still an open research area in linguistics \cite{lester2017you,blaszczak2019investigation}. 

Lexical richness and complexity relate to nuances and the intricacy of meaning in language. Numerous metrics to quantify lexical diversity, such as type-token ratio (TTR)~\cite{lex_richards1987type} and MLTD~\cite{lex_mccarthy2005assessment}, have been proposed. These metrics capture various dimensions of meaning, quantity and quality 
of words, such as variability, volume, and rarity. Several of these have been identified to be important for a variety of tasks in applied linguistics~\cite{lex_daller2003lexical}. 
For example, metrics related to vocabulary size, such as TTR and word-frequencies, have proven to help with early detection 
of mild cognitive impairment (MCI)~\cite{lex_aramaki2016vocabulary}, hence are important for early dementia diagnosis. Discourse informativeness, measured via propositional idea density, is also shown to be significantly affected in speakers with aphasia ~\cite{lex_bryant2013propositional}. Furthermore, lexicon-based methods have proved to be successful in sentiment analysis~\cite{lex_taboada2011lexicon, lex_tang2014learning}.

Syntactic complexity is evident in language production in terms of syntactic variation and sophistication or, in other words, the range and degree of sophistication of the syntactic structures that are produced \cite{lu2011corpus,ortega2003syntactic}. This construct has attracted attention in a variety of language-related research areas. For example, researchers have examined the developmental trends of child syntactic acquisition (e.g.,~\cite{langacq_ramer1977development}), the role of syntactic complexity in treating syntactic deficits in agrammatical aphasia (e.g.,~\cite{aph_melnick2000relationship,aph_thompson2003role}), the relationship between syntactic complexity in early life to symptoms of Alzheimer’s disease in old age (e.g.,~\cite{ad_kemper2001longitudinal,ad_snowdon1996linguistic}), and the effectiveness of syntactic complexity as a predictor of adolescent writing quality (e.g.,~\cite{syn_beers2009syntactic}).

\citet{indefrey2001syntactic} reported data on brain activation during syntactic processing and demonstrated that syntactic processing in the human brain happens independently of the processing of lexical meaning. These results were supported by the more recent studies showing that different brain regions support distinct mechanisms in the mapping from a linguistic form onto meaning, thereby separating syntactic agrammaticality from linguistic complexity\cite{ullman2005neural,friederici2006processing}. This motivates us to explore the importance of lexical and syntactic features separately. 

To our knowledge, there is no previous research in medical text classification area exploring the individual value of lexical and syntactic features with regards to their vulnerability and importance for ML models. Syntactic and lexical feature groups are often used together without specifying their individual value. For example, recent work in text classification for AD detection revealed that a combination of lexical and syntactic features works well~\cite{ad_fraser2016detecting, ad_noorian2017importance}; the same is true for other cognitive disease or language impairment detection~\cite{aph_meteyard2009relation, aph_fraser2014automated}, as well as sentiment detection in healthy speech and language~\cite{sent_negi2014insight,sent_marchand2013lvic,sent_pang2002thumbs}. 

In this paper, we focus on individual value of lexical and syntactic feature groups, as studied across medical text classification tasks, types of language, datasets and domains. As such, the main contributions of this paper are: 

\begin{itemize}
\item Inspired by the results of neuroscience studies~\cite{indefrey2001syntactic}, we explore selective performance of lexical and syntactic feature groups separately.

\item We demonstrate, using multiple analysis methods, that there is a clear difference in how lexical features endure text alterations in comparison to the syntactic ones as well as how the latter impact classification.

\item We report results on three different datasets and four different classifiers, which allows us to draw more general conclusions.

\item We conduct an example-based analysis that explains the results obtained during the analysis.
\end{itemize}

\section{Related Work}
\label{sec:related_work}

Prior research reports the utility of different modalities of speech -- lexical and syntactic~\cite{bucks2000analysis, ad_fraser2016detecting,ad_noorian2017importance,zhu2018detecting} -- in detecting dementia.~\citet{bucks2000analysis} obtained a cross-validated accuracy of 87.5\% among a sample of 24 participants in detecting AD using eight linguistic features, including part-of-speech (POS) tag frequencies and measures of lexical diversity. A similar feature set was employed by ~\citet{ad_meilan2014speech} in a larger sample, where measures  of lexical richness were less useful than features indicative of word finding difficulty (such as pauses and repetitions).~\citet{ad_orimaye2014learning} obtained F-measure scores up to 0.74 using a combination of lexical and syntactic features on transcripts from a large dataset of AD and controls speech, DementiaBank (see Section~\ref{sec:datasets})

Similarly, varying feature sets have been used for detecting aphasia from speech. Researchers have studied the importance of syntactic complexity indicators such as Yngve-depth and length of various syntactic representations for detecting aphasia~\cite{aph_roark2011spoken}, as well as lexical characteristics such as average frequency and the imageability of words used~\cite{aph_bird2000rise}. Patterns in production of nouns and verbs are also particularly important in aphasia detection~\cite{aph_wilson2010connected, aph_meteyard2009relation}. ~\citet{aph_fraser2014automated} used a combination of syntactic and lexical features with ASR-transcription for the diagnosis of primary progressive aphasia with a cross-validated accuracy of 100\% within a dataset of 30 English-speakers. More recently, ~\citet{aph_le2017automatic} proposed methods to detect paraphasia, a type of language output error commonly associated with aphasia, in aphasic speech using phone-level features.

Sentiment analysis methodologies often use lexicon-based features~\cite{lex_taboada2011lexicon, lex_tang2014learning}. Syntactic characteristics of text such as proportions of verbs and adjectives, nature of specific clauses in sentences are also salient in sentiment detection~\cite{sent_chesley2006using, sent_meena2007sentence}. Additionally, systems using both syntactic and lexical features have been proposed in prior work~\cite{sent_negi2014insight,sent_marchand2013lvic}. For example, ~\citet{sent_marchand2013lvic} trained ML models on patterns in syntactic parse-trees and occurrences of words from a sentiment lexicon to detect underlying sentiments from tweets while ~\citet{sent_negi2014insight} employed syntactic and lexical  features for sentence level aspect based sentiment analysis.~\citet{sent_pang2002thumbs} showed that unigrams, bigrams and frequencies of parts-of-speech tags such as verbs and adjectives are important for an ML-based sentiment classifier. 

\section{Method}

\begin{table}[t]
\begin{center}
\begin{adjustbox}{max width=0.49\textwidth}
\begin{tabular}{ll|c|c|c}
 &  & \multicolumn{3}{c}{\textbf{Datasets}} \\
 &  & \textbf{DemB} & \textbf{IMDBs} & \textbf{AphB} \\
\hline 
\hline
\multirow{2}{*}{Task nature} & Structured & X & X &  \\
\cdashline{2-5}[0.5pt/2pt]
 & Partially structured &  &  & X \\
\hline
\multirow{2}{*}{Language type} & Verbal & X &  & X \\
\cdashline{2-5}[0.5pt/2pt]
 & Written &  & X &  \\
\hline
\multirow{3}{*}{Lexics} & Complex &  &  & X \\
\cdashline{2-5}[0.5pt/2pt]
 & Medium &  & X &  \\
\cdashline{2-5}[0.5pt/2pt]
 & Simple & X &  &  \\
\hline
\multirow{3}{*}{Syntax} & Complex &  & X &  \\
\cdashline{2-5}[0.5pt/2pt]
 & Medium & X &  &  \\
\cdashline{2-5}[0.5pt/2pt]
 & Simple &  &  & X \\
\hline
\end{tabular}
\end{adjustbox}
\caption{Comparison of the datasets in terms of task nature, type of language used to collect the data, lexical and syntactic complexity.}
\label{tab:ds_compare}
\end{center}
\end{table}

\subsection{Datasets}
\label{sec:datasets}
In  the  following  section,  we  provide  context on each of three similarly-sized datasets that we investigate that differ in the following ways (see also Section~\ref{sec:ds-comparison}):
\begin{enumerate}
    \itemsep 0em
    \item Binary text classification task (AD detection, sentiment classification, aphasia detection).
    \item Type of language
    \item Level of lexical and syntactic complexity.
\end{enumerate}

\subsubsection{DementiaBank (DemB)}
DementiaBank\footnote{https://dementia.talkbank.org} is 
the largest publicly available dataset for detecting cognitive impairments, and is a part of the TalkBank corpus ~\cite{macwhinney2007talkbank}. It consists of 
audio recordings of  verbal descriptions and associated transcripts of the Cookie Theft picture description task from the Boston Diagnostic Aphasia Examination \cite{ad_becker1994natural} from 210 participants aged between 45 to 90. Of these participants, 117 have a clinical diagnosis of AD ($N=180$ speech recordings), while 93 ($N=229$ speech recordings) are cognitively healthy. Many participants repeat the task within an interval of a year. 

\subsubsection{AphasiaBank (AphB)}
AphasiaBank\footnote{https://aphasia.talkbank.org}~\cite{macwhinney2007talkbank} is another dataset of pathological speech that consists of aphasic and healthy control speakers performing a set of standard clinical speech-based tasks. The dataset includes audio samples of speech and associated transcripts. All participants perform multiple tasks, such as describing pictures, story-telling, free speech, and discourse with a fixed protocol. Aphasic speakers have various sub-types of aphasia (fluent, non-fluent, etc.). In total, there are 674 samples, from 192 healthy ($N=246$ speech samples) and 301 ($N=428$ speech samples) aphasic speakers.

\subsubsection{IMDB Sentiment Extract (IMDBs)}
The IMDB Sentiment~\cite{sent_maas2011learning} dataset is a standard corpus for sentiment detection that contains typewritten reviews of movies from the IMDB database along with the review-associated binary sentiment polarity labels (positive and negative). This dataset is used in order to extend the range of `healthy' language and test generalizability of our findings. The core dataset consists of 50,000 reviews split evenly into train and test sets (with equal classes in both train and test). To maintain a comparable dataset size to DemB and AphB, we randomly choose 250 samples from the train sets of each polarity, totalling 500 labeled samples.

All the three datasets cover a breadth of transcripts in terms of presence or absence of impairment,  as well as a spectrum of `healthy' speech.


\subsection{Feature Extraction}

Following multiple previous works on text classification, we extract two groups of linguistic features -- lexical and syntactic.

\paragraph{Lexical features:}
 
Features of lexical domain have been recognized as an important construct in a number of research areas, including stylistics, text readability analysis, language assessment, first and second language acquisition, and cognitive disease detection. In order to measure various dimensions of lexical richness in the datasets under comparison, we compute statistics on token/unigram, bigram, and trigram counts. Additionally, we use the Lexical Complexity Analyser \cite{ai2010web} 
to measure various dimensions of lexical richness, such as lexical density, sophistication, and variation.

Following \citet{oraby_controlling_2018}, \citet{duvsek2019evaluating}, and \citet{jagfeld_sequence--sequence_2018}, we also use Shannon entropy \cite[p.~61ff.]{manning_foundations_2000} as a measure of lexical diversity in the texts:

\begin{dmath}
H(\mbox{text}) = - \sum_{x\,\in\,\mbox{text}} \frac{\mbox{freq}(x)}{\mbox{len}(\mbox{text})} \log_2 \left(\frac{\mbox{freq}(x)}{\mbox{len}(\mbox{text})}\right)
\end{dmath}

Here, $x$ stands for all unique tokens/$n$-grams, \emph{freq} stands for the number of occurrences in the text, and \emph{len} for the total number of tokens/n-grams in the text. We compute entropy over tokens (unigrams), bigrams, and trigrams.

We further complement Shannon text entropy with $n$-gram conditional entropy for next-word prediction \cite[p.~63ff.]{manning_foundations_2000}, given one previous word (bigram) or two previous words (trigram):

\vspace{0.3em}
\begin{dmath}
H_{cond}(\mbox{text}) = - \sum_{(c,w)\,\in\,\mbox{text}} \frac{\mbox{freq}(c,w)}{\mbox{len}(\mbox{text})} \log_2 \left(\frac{\mbox{freq}(c,w)}{\mbox{freq}(c)}\right)
\end{dmath}
\vspace{0.3em}

Here, $(c,w)$ stands for all unique $n$-grams in the text, composed of $c$ (context, all tokens but the last one) and $w$ (the last token).
Conditional next-word entropy gives an additional, novel measure of diversity and repetitiveness: the more diverse text is, 
the less predictable is the next word given the previous word(s) is; on the other hand, the more repetitive the text, 
the more predictable is the next word given the previous word(s).

\begin{table}[t]
\begin{center}
\begin{adjustbox}{max width=0.495\textwidth}
\begin{tabular}{ll|ccc}
\begin{tabular}[c]{@{}l@{}}\textbf{Feature} \\ \textbf{subgroup}\end{tabular} & \textbf{Feature} & \textbf{DemB} & \textbf{IMDBs} & \textbf{AphB} \\
\hline 
\hline
\multirow{3}{*}{\begin{tabular}[c]{@{}l@{}}Lexical\\ richness\end{tabular}} & distinct tokens occuring once, \% & 0.58 & \textbf{0.64} & 0.32 \\
 & distinct bigrams occuring once, \% & 0.89 & \textbf{0.95} & 0.83 \\
 & distinct trigrams occuring once, \% & 0.96 & \textbf{0.99} & 0.92 \\
 \cdashline{1-5}[0.5pt/2pt]
\multirow{6}{*}{\begin{tabular}[c]{@{}l@{}}Lexical\\ complexity\end{tabular}} & unigram entropy & 5.42 & 6.53 & \textbf{6.70} \\
 & bigram entropy & 6.4 & 7.46 & \textbf{8.68} \\
 & trigram entropy & 6.55 & 7.54 & \textbf{9.19} \\
 & bigram conditional entropy & 1.01 & 0.95 & \textbf{1.99} \\
 & trigram conditional entropy & 0.16 & 0.09 & \textbf{0.51} \\
 & lexicon complexity & 1.33 & \textbf{1.47} & 1.32 \\
\hline
\multirow{3}{*}{\begin{tabular}[c]{@{}l@{}}Length of\\ production\\ unit\end{tabular}} & Mean length of clause & 7.45 & \textbf{9.24} & 5.42 \\
 & Mean length of sentence & 8.77 & \textbf{21.42} & 6.01 \\
 & Mean length of T-unit & 11.85 & \textbf{18.99} & 6.15 \\
 \cdashline{1-5}[0.5pt/2pt]
\multirow{4}{*}{\begin{tabular}[c]{@{}l@{}}Sentence\\ complexity\end{tabular}} & Clauses per sentence & 1.21 & \textbf{2.35} & 1.08 \\
 & D-level 0 & 0.63 & 0.26 & \textbf{0.74} \\
 & D-level 1-4 & \textbf{0.23} & 0.21 & 0.14 \\
 & D-level 5-7 & 0.14 & \textbf{0.52} & 0.11 \\
 \cdashline{1-5}[0.5pt/2pt]
\multirow{4}{*}{\begin{tabular}[c]{@{}l@{}}Amount of\\ subordination\end{tabular}} & Clauses per T-unit & 1.62 & \textbf{2.07} & 1.12 \\
 & Complex T-units per T-unit & 0.19 & \textbf{0.55} & 0.14 \\
 & Dependent clauses per T-unit & 0.68 & \textbf{1.00} & 0.19 \\
 \cdashline{1-5}[0.5pt/2pt]
\multirow{3}{*}{\begin{tabular}[c]{@{}l@{}}Amount of\\ coordination\end{tabular}} & Coordinate phrases per clause & 0.11 & \textbf{0.22} & 0.10 \\
 & Coordinate phrases per T-unit & 0.17 & \textbf{0.44} & 0.11 \\
 & T-units per sentence & 0.77 & \textbf{1.13} & 0.95 \\
 \cdashline{1-5}[0.5pt/2pt]
\multirow{3}{*}{\begin{tabular}[c]{@{}l@{}}Particular\\ structures\end{tabular}} & Complex nominals per clause & 0.64 & 1.09 & 0.33 \\
 & Complex nominals per T-unit & 1.03 & \textbf{2.28} & 0.38 \\
 & Verb phrases per T-unit & 1.93 & \textbf{2.64} & 1.19\\
 \hline
 \end{tabular}
\end{adjustbox}
\end{center}
\caption{Lexical complexity and richness, and syntactic complexity of the three datasets. Counts for $n$-grams appearing only once are shown as proportions of the total number of respective $n$-grams. Highest values on each line are typeset in bold.}
\label{tab:ds_overview}
\end{table}

\paragraph{Syntactic Features:}

We used the D-Level Analyser \cite{lu2009automatic} to evaluate syntactic variation and complexity of human references using the revised D-Level Scale \cite{lu2014computational}.

We use the L2 Syntactic Complexity Analyzer \cite{lu2010automatic} to extract 14 features of syntactic complexity that represent the length of production units, sentence complexity, the amount of subordination and coordination, and the frequency of particular syntactic structures. The full list of lexical and syntactic features is provided in Appendix~\ref{app:feature_list}.

\subsection{Classification Models}
We benchmark four different machine learning models on each dataset with 10-fold cross-validation. In cases of multiple samples per participant, we stratify by subject so that samples of the same participant do not occur in both the train and test sets in each fold. This is repeated for each text alteration level. The minority class is oversampled in the training set using SMOTE~\cite{chawla2002smote} to deal with class imbalance. 

We consider Gaussian na\"ive Bayes (with equal priors), random forest (with 100 estimators and 
maximum depth 5), support vector Machine (with RBF kernel, penalty $C=1$), and a 2-hidden layer neural network (with 10 units in each layer, ReLU activation, 200 epochs and Adam 
optimizer)~\cite{pedregosa2011scikit}. Since the datasets have imbalanced classes, we identify F1 score with macro averaging as the primary performance metric.

\subsection{Altering Text Samples}

There can be three types of language perturbations at the word level: insertions, deletions, and substitutions on words. \cite{balagopalan2019impact} showed that deletions are more affected (significantly) than insertions and substitutions, so we likewise focus on deletions. Following \citet{balagopalan2019impact}, we artificially add deletion errors to original individual text samples at predefined levels of 20\%, 40\%, 60\%, and 80\%. To add the errors, we simply delete random words from original texts and transcripts at a specified rate.

\subsection{Evaluating Change of Feature Values}
\label{sec:feature_delta}
In order to evaluate the change of feature values for different levels of text alterations, $z$-scores are used. We calculate $z$-scores of each individual feature in the transcripts with each level of alteration, with relation to the value of that feature in the original unaltered transcript. 

\vspace{0.3em}
\begin{dmath}
    Z_{feat}^{x}= (feat^{x}-\mu_{no-alteration})/\sigma_{no-alteration},
\end{dmath}
\vspace{0.5em}

where $feat^{x}$ refers to a given syntactic or lexical feature extracted from a transcript with an alteration level of $x=20..80$, $\mu$ and $\sigma$ are computed over the entire original unaltered dataset.

Then, we average the individual $z$-scores across all the features within each feature group (syntactic and lexical) to get a $z$-score per feature group.
\vspace{0.3em}
\begin{equation}
    Z_{syntactic}^{x} = \frac{1}{N_{syn}}\sum_{i=1}^{N_{syn}}Z_{feat_{i}}^{x}
\end{equation}

\begin{equation}
    Z_{lexical}^{x} = \frac{1}{N_{lex}}\sum_{i=1}^{N_{lex}}Z_{feat_{i}}^{x},
\end{equation}
\vspace{0.3em}

where $N_{syn}$ and $N_{lex}$ refer to the total number of syntactic and lexical features, respectively.

\subsection{Evaluating change of feature predictive power}
\label{sec:pred-power}
We extract $\Delta F1_{x}$, or change in classification F1 macro score, with $x\%$ alteration with respect to no alteration, for $x=20,40,60,80$, i.e, 

\vspace{-0.3em}
\begin{equation}
    \Delta F1_{x} = F1_{x\%alteration} - F1_{no-alteration}.
\end{equation}

To identify the relative importance of syntactic or lexical features on classification performance, we estimate coefficients of effect for syntactic and lexical features. These coefficients are obtained by regressing to F1 deltas using the syntactic and lexical feature $z$-scores described in Section~\ref{sec:feature_delta} for each alteration level. Thus, the regression equation can be expressed as:

\begin{equation}
  \Delta F1 = \alpha Z_{syntactic} + \beta Z_{lexical}.
\end{equation}

The training set for estimating $\alpha$ and $\beta$ consists of {$\Delta F1_{x}; (Z_{syntactic}^{x},Z_{lexical}^{x})$} for $x=20, 40, 60, 80$.

\section{Comparing datasets}
\label{sec:ds-comparison}

Three datasets used in our exploration represent different dimensions of lexical and syntactic complexity, and are unique in the nature of the tasks they involve and their type of language, as shown in Tab.\ref{tab:ds_compare}. AphB is the only dataset that includes speech samples of unstructured speech, while IMDBs is unique as it contains samples of written language, rather than transcripts of verbal speech.

In terms of lexical and syntactic complexity, it is interesting to note that AphB contains samples that are most lexically complex, while at the same time it is the most simple from the syntactic point of view. We associate this with the fact that AphB data come from partially unstructured tasks, where free speech increases the use of a more complex and more diverse vocabulary. 
IMDB is the most lexically rich dataset (see Table~\ref{tab:ds_overview}), with the highest ratio of uni-, bi-, and trigrams occuring only once.

IMDB is the most complex according to various measures of syntactic complexity: 
it has the highest scores with metrics associated with length of production unit, amount of subordination, coordination, and particular structures, and it also has the highest amount of complex sentences (sentences of D-level 5-7, as shown in Table~\ref{tab:ds_overview}). This may be explained by the fact it is the only dataset based on typewritten language. AphB has the lowest level of syntactic complexity, containing the highest amount of the simplest sentences (D-level 0), and lowest scores in other subgroups of syntactic features (see Table~\ref{tab:ds_overview}).

Next, we analyse if these variously distinct datasets have any common trend with regards to the vulnerability and robustness of lexical and syntactic feature groups.

\section{Results and discussion}

\begin{table}[tb]
\begin{center}
\begin{adjustbox}{max width=0.49\textwidth}
\begin{tabular}{lc|cc}
\textbf{Dataset} & \begin{tabular}[c]{@{}l@{}}\textbf{Level (\%) of} \\ \textbf{alterations}\end{tabular} & \begin{tabular}[c]{@{}l@{}}\textbf{Lexical}\\ \textbf{features (z-score)}\end{tabular} & \begin{tabular}[c]{@{}l@{}}\textbf{Syntactic} \\ \textbf{features (z-score)}\end{tabular} \\
\hline \hline
\multirow{4}{*}{DemB} & 20 & \textbf{0.35} & 0.30 \\
 & 40 & \textbf{0.75} & 0.62 \\
 & 60 & \textbf{1.21} & 0.94 \\
 & 80 & \textbf{1.85} & 1.31 \\
 \cdashline{1-4}[0.5pt/2pt]
\multirow{4}{*}{AphB} & 20 & 0.10 & \textbf{0.15} \\
 & 40 & \textbf{0.26} & 0.26 \\
 & 60 & \textbf{0.51} & 0.37 \\
 & 80 & \textbf{0.93} & 0.51 \\
 \cdashline{1-4}[0.5pt/2pt]
\multirow{4}{*}{IMDBs} & 20 & \textbf{0.29} & 0.13 \\
 & 40 & \textbf{0.61} & 0.25 \\
 & 60 & \textbf{1.00} & 0.35 \\
 & 80 & \textbf{1.61} & 0.31 \\
 \hline
\end{tabular}
\end{adjustbox}
\end{center}
\caption{Change of feature values, per dataset and per level of text alterations.}
\label{tab:deltas}
\end{table}

\begin{figure*}[t!]
\centering
    \begin{subfigure}[b]{0.452\textwidth}
        \includegraphics[width=\textwidth]{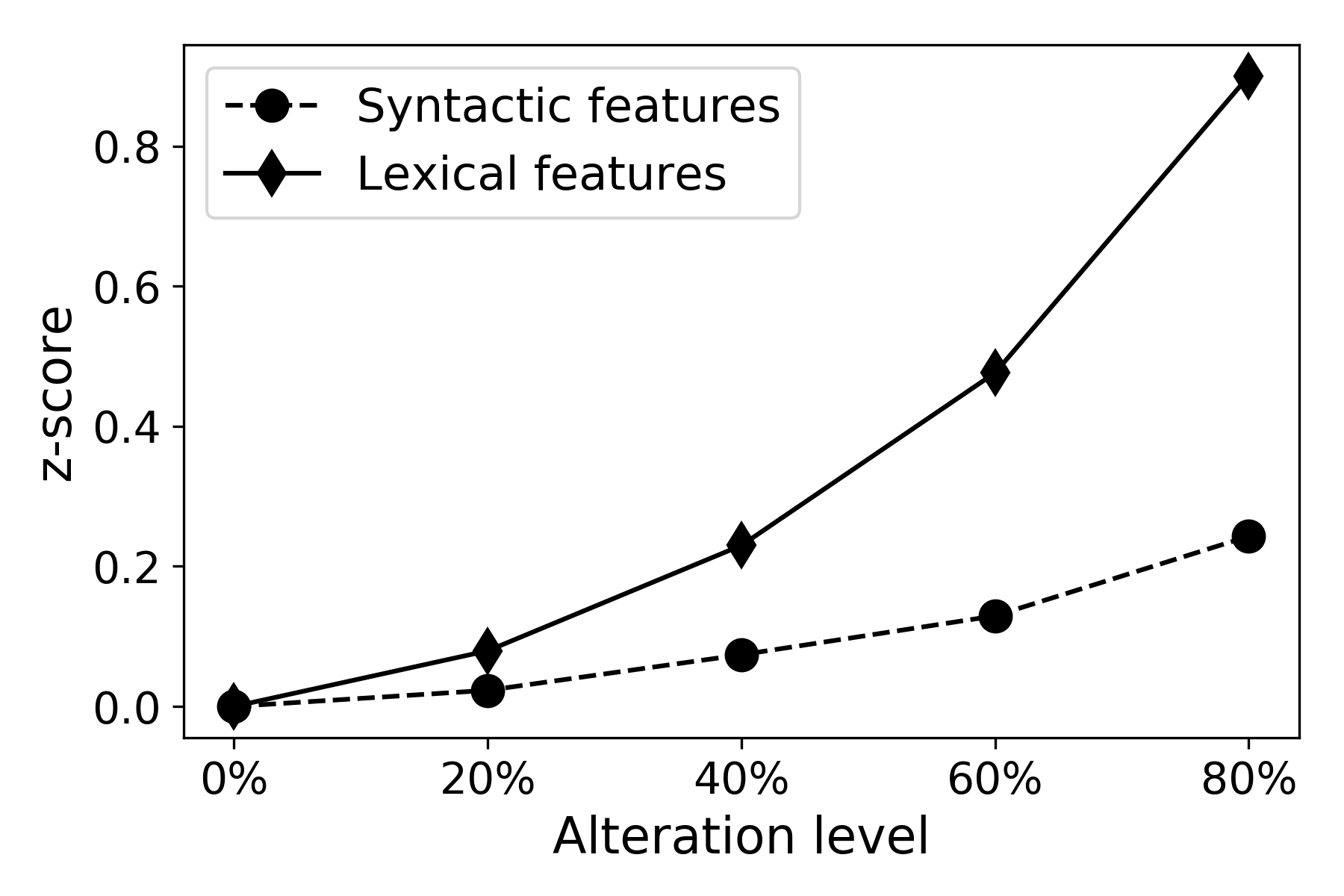}
    \end{subfigure} 
    \begin{subfigure}[b]{0.452\textwidth}
        \includegraphics[width=\textwidth]{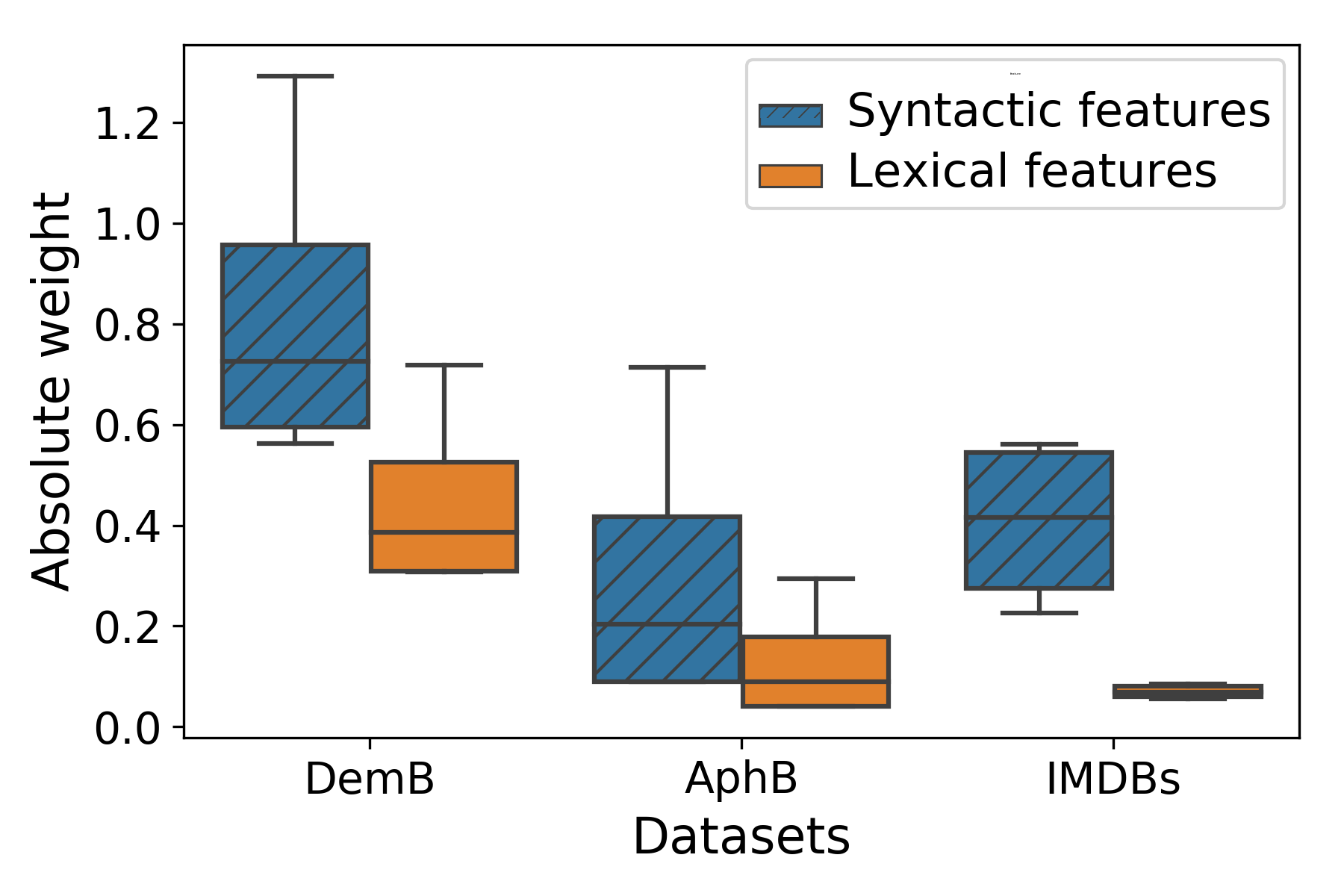}
    \end{subfigure}
\caption{Left: Change of syntactic and lexical feature values at different alteration levels, averaged across three datasets. Right: Impact of syntactic and lexical features on classification for DementiaBank, AphasiaBank and IMDBsentiment datasets, averaged across fours classifiers.}
\label{fig:ngrams}
\end{figure*}

\subsection{Feature vulnerability}

Following the method described in Section~\ref{sec:feature_delta}, we analyse if any of the feature groups (lexical or syntactic) is influenced more by text alterations. As shown in Figure~\ref{fig:ngrams}, the values of lexical features are, on average, influenced significantly more than syntactic ones (Kruskal-Wallis test, $p<$0.05). Such a difference is observed in all three datasets individually (see Table~\ref{tab:deltas}). 

The differences of $z$-scores between lexical and syntactic feature groups are higher for the IMDBs dataset, which suggests that the difference is most visible either in healthy or in written language. 

These results suggest that lexical features are more vulnerable to simple text alterations, such as introduced deletion errors, while syntax-related features are more robust to these modifications. However, stronger changes of raw feature values do not necessarily mean that the resulting modified features become more or less important for classifiers. This leads us to inspect the impact of text alteration on feature predictive power.

\subsection{Feature significance and the impact of alterations on feature predictive power}

\begin{figure*}[t!]
\centering
    \begin{subfigure}[b]{0.408\textwidth}
        \includegraphics[width=\textwidth]{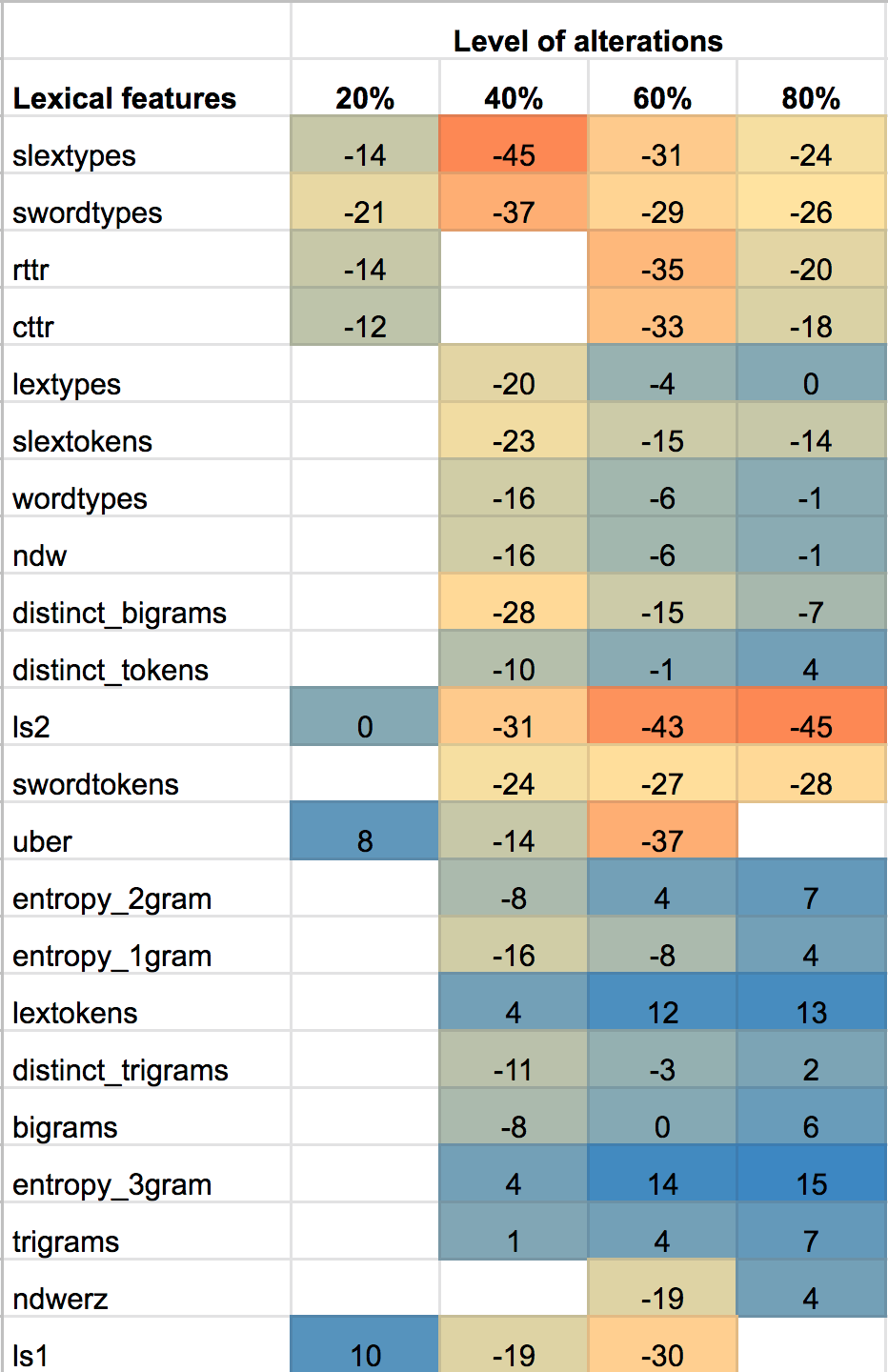}
    \end{subfigure} 
    \begin{subfigure}[b]{0.47\textwidth}
        \includegraphics[width=\textwidth]{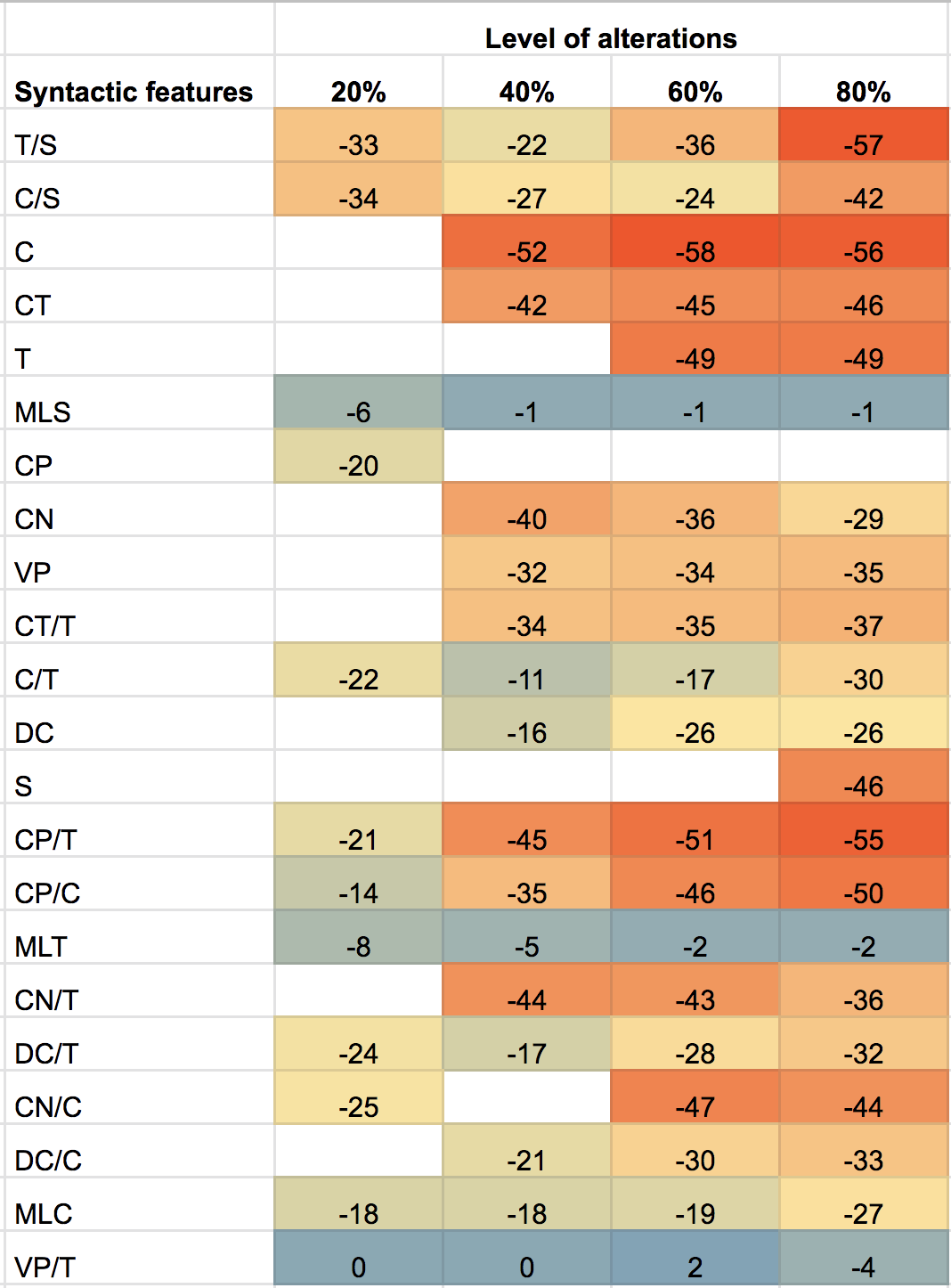}
    \end{subfigure}
\caption{Change of lexical (left) and syntactic (right) feature rank when text alterations of different levels are introduced. Negative numbers denote decrease in rank, and positive numbers are an increase of rank. Blue cell colours denote the highest increase in rank, red (the highest decrease) and yellow (a smaller level of increase or decrease). Features are ranked based on p-values with the lowest p-value at the top. White cells show that features were not significantly different between classes in the original text samples, based on DemB dataset.}
\label{fig:ranks}
\end{figure*}

A simple method to understand the potential predictability of a feature is by looking at how different the feature value is between classes and whether this difference is statistically significant. This method was previously used in studies assessing automatic speech recognition for Alzheimer's \cite{zhou2016speech} and aphasia detection \cite{fraser2013automatic}. 

We rank the $p$-values obtained, in each condition, from a two tailed non-parametric Kruskal-Wallis test performed on each feature between the two classes (healthy vs unhealthy in the DB and AphB datasets, and positive vs negative in IMDBs) and assign rank to each feature. It is interesting to note that lexical features occupy the overwhelming majority of first places across all datasets, showing that lexical features are significantly different between classes. We further analyse, following \cite{brunato2018sentence}, how the rank of each feature changes when different levels of text alterations are introduced. The maximum rank increase is higher on average for lexical features than for syntactic (see Figure~\ref{fig:ranks} for details of rank changes in DemB dataset) across all datasets. The ratio of features that become insignificant after text alteration is also higher for lexical features rather than in syntactic on average across all datasets. As Figure~\ref{fig:ranks} shows, the features with increased rank are those that were not initially significantly different between classes. The combination of these results suggest that not so important lexical features become more and more important with addition of text alterations, which may decrease the performance of classification. 

The above method of calculating $p$-values is analogous to feature selection performed as a pre-processing step before classification. Although this step may provide some initial insights into feature significance, it does not guarantee the most significant features will be those having the most predictive power in classification.

\begin{table}[t!]
\begin{center}
\begin{adjustbox}{max width=0.40\textwidth}
\begin{tabular}{l|c|c|c|c}
Dataset & \multicolumn{3}{c}{\textbf{Classifiers}} \\
& \textbf{NN} & \textbf{SVM} & \textbf{RF}& \textbf{NB} \\
 \hline \hline
DemB & 1.82	& 1.83 & 1.98 & 1.80\\
IMDBs & 5.22 & 6.39 & 7.15 & 3.74\\
AphB & 2.22	& 2.44 & 2.28 & 2.17 \\

 \hline
\end{tabular}
\end{adjustbox}
\caption{Ratio of coefficients, calculated as $Importance_{syntactic} / Importance_{lexical}$. Ratio higher than one indicates that syntactic features are more important for a classifier than lexical ones.}
\label{tab:syn_lex_ratio}
\end{center}
\end{table}

We use the method described in Section~\ref{sec:pred-power} to evaluate the impact of text alteration on the feature’s predictive power. The results in Table~\ref{tab:syn_lex_ratio} show that syntactic features have more predictive power than lexical features. The lowest ratio is observed with DemB, and the AphB results are very close, suggesting that syntactic features are approximately twice as important than lexical features in predicting pathological speech. In healthy written language, the difference is even higher and reaches 7.15 for the random forest classifier. 

In summary, the predictive power of syntactic features is much stronger than that of lexical features across three datasets and four main classifiers, which suggest the results can be generalizable across several different tasks and domains.


\subsection{Example-based Analysis}

As shown in previous sections, values of lexical features are on average more influenced by text alterations but this change does not affect classification as much as smaller value changes in syntactic features. Table~\ref{tab:example} provides examples of two features, one lexical and one syntactic, their value changes when text samples are modified, and the associated change of the classifier's predictions.

\begin{table*}[]
\begin{center}
\begin{adjustbox}{max width=0.99\textwidth}
\begin{tabular}{lm{20cm}llll}
\textbf{\begin{tabular}[c]{@{}l@{}}Alteration\\ level\end{tabular}} & \textbf{Text sample} & \textbf{Feature} & \textbf{\begin{tabular}[c]{@{}l@{}}Feature\\ value / $\Delta$\end{tabular}} & \textbf{Prediction} & \textbf{Dataset} \\
\hline \hline
original & \&uh the boy is reaching into the cookie jar. he's falling off the stool. the little girl is reaching for a cookie. mother is drying the dishes. the sink is running over. mother's getting her feet wet. they all have shoes on. there's a cup two cups and a saucer on the sink. the window has draw withdrawn drapes. you look out on the driveway. there's kitchen cabinets. oh what's happening. mother is looking out the window. the girl is touching her lips. the boy is standing on his right foot. his left foot is sort of up in the air. mother's right foot is flat on the floor and her left she's on her left toe. \&uh she's holding the dish cloth in her right hand and the plate she is drying in her left. I think I've run out of. yeah. & \begin{tabular}[c]{@{}l@{}}lexical\\ (cond\_entropy\_3gram)\end{tabular} & 0.24 / - & Correct (healthy) & DemB \\
 \hline
20\% & \&uh the boy reaching the cookie jar. he's falling off the stool. the little girl is reaching for cookie. mother is the dishes. the sink is over. mother's getting her feet. all have shoes. there's cup two cups a saucer on sink. window has draw withdrawn drapes. you look out on driveway. there's kitchen cabinets. oh what's happening. mother out the window. the girl is lips. the boy standing on. his left foot is sort of up in the air. mother's right foot is flat on the floor and left she's on her left toe. \&uh she's holding the cloth in right hand the plate she drying in her left. think I've run out of. & \begin{tabular}[c]{@{}l@{}}lexical\\ (cond\_entropy\_3gram)\end{tabular} & 0.11 / 0.48 & Correct (healthy) & DemB \\
 \hline
40\% & \&uh reaching the jar. he's falling the stool. the little is reaching a cookie. mother drying the dishes. the sink is running over. mother's her wet. all have shoes on. a two and a sink. the. you look driveway. there's kitchen. oh what's happening. mother out the window. the is her. is his foot. his left foot is sort of up air. foot is flat floor and she's her toe. \&uh she's holding the dish cloth in right the she is drying in left. I think of. & \begin{tabular}[c]{@{}l@{}}lexical\\ (cond\_entropy\_3gram)\end{tabular} & 0.07 / 0.28 & Correct (healthy) & DemB \\
 \hline
60\% & \&uh is cookie. falling stool. for cookie. the dishes. the. mother's feet wet. they have. a two cups a sink. the has withdrawn drapes. the. there's. oh. mother the window. the lips. the boy right. is sort of. right foot is flat on floor on her left. \&uh cloth right hand and the she is in her left. yeah. & \begin{tabular}[c]{@{}l@{}}lexical\\ (cond\_entropy\_3gram)\end{tabular} & 0.03 / 0.14 & Incorrect (AD) & DemB \\
 \hline \hline
 original & okay. well in the first place the the mother forgot to turn off the water and the water's running out the sink. and she's standing there. it's falling on the floor. the child is got a stool and reaching up into the cookie jar. and the stool is tipping over. and he's sorta put down the plates. and she's reaching up to get it but I don't see anything wrong with her though. yeah that's it. I can't see anything. & syntactic (C/S) & 1.1 / - & Correct (healthy) & DemB \\
 \hline
20\% & well the first the the mother forgot to turn off the water the water's out the sink. and standing there. it's falling floor. is got a stool and into the cookie jar. and the stool is tipping. and he's sorta down the plates. and she's reaching to get it but I don't see anything wrong with her though. that's it. I can't see anything. & syntactic (C/S) & 1.22 / 1.11 & Incorrect (AD) & DemB \\
 \hline
40\% & okay. well in the forgot the water the water's out the sink. and she's standing there. it's on the. the is got a stool and reaching up the. the is tipping. and he's sorta the. and she's reaching up to get but I her. yeah that's. I can't. & syntactic (C/S) & 1.0 / 0.91 & Incorrect (AD) & DemB \\
 \hline
60\% & okay. in water's out the sink. falling. the got stool the cookie jar. and the stool is over. and he's down the plates. and she's up but don't wrong. can't see anything. & syntactic (C/S) & 1.0 / 0.91 & Incorrect (AD) & DemB \\
\hline
\end{tabular}
\end{adjustbox}
\caption{Examples of two features, \textit{cond\_entropy\_3gram} and \textit{C/S}, their value change when text samples are modified on the level of 20\%, 40\% and 60\%, and associated classifier's predictions. Examples are provided using the DemB transcript samples and feature values.\label{tab:example}}
\end{center}
\end{table*}

The value of lexical feature \textit{cond\_entropy\_3gram}, showing conditional entropy calculated for trigrams, decreases by more than 50\% when the text sample is modified by only 20\%. This change is much higher than the associated absolute change of the syntactic feature \textit{C/S} (that shows the number of clauses per sentence) that increases by 11\% only on the same level of alteration. The prediction made by a classifier in the case of the lexical feature, however, is the same as the prediction of original transcript. Only when the general level of alteration reaches 60\% and the value of the lexical feature decreases by more than 85\%, the prediction becomes incorrect. In the case of syntactic features, the prediction already changes to incorrect with the general level of alteration of 20\%, although the feature value is still quite close to the original one.

Consider this sentence in the original transcript: 
\vspace{0.5em}
\begin{addmargin}[2em]{1em}
\textit{She's holding the dish cloth in her right hand and the plate she is drying in her left}.
\end{addmargin}
With 20\% of errors it is converted to the following: 
\begin{addmargin}[2em]{1em}
\textit{She's holding the cloth in right hand the plate she drying in her left.}
\end{addmargin}
It is clear that lexical features based on the frequency of uni-, bi- and trigrams are affected by this change, because quite a few words disappear in the second variant. In terms of syntactic structures, however, the sentence is not damaged much, as we still can see the same number of clauses, coordinate units, or verb phrases. Such an example helps explain the results in the previous sections.

\section{Conclusions and Future Research}
\label{sec:concl}

This paper shows that linguistic features of text, associated with syntactic and lexical complexity, are not equal in their vulnerability levels, nor in their predictive power. We study selective performance of these two feature aggregations on three distinct datasets to verify the generalizability of observations. 

We demonstrate that values of lexical features are easily affected by even slight changes in text, by analysing $z$-scores at multiple alteration levels. Syntactic features, however, are more robust to such modifications. On the other hand, lower changes of syntactic features result in stronger effects on classification performance. Note that these patterns are consistently observed across different datasets with different levels of lexical and syntactic complexity, and for typewritten text and transcribed speech. 

Several methods to detect and correct syntactic~\cite{ma2012detecting} and lexical errors~\cite{klebesits1994lexical} as a post-processing step for output from machine translation or ASR systems have been proposed in prior work.
Since our analysis indicates that error-affected syntactic features have a stronger effect on classification performance, we suggest imposing higher penalties on detecting and correcting syntactic errors than lexical errors in medical texts. A limitation in our study is that we focused on text alterations of a specific type, and the results were only tested on relatively small datasets. In future work, we will extend the analysis to other simple text alterations such as substitutions as well as adversarial text attacks~\cite{alzantot2018generating}. 
In addition, we will extend the current work to see how state-of-the-art neural network models, such as Bert, can handle text alterations as they capture lexical, syntactic and semantic features of the input text in different layers.
Finally, note that the datasets considered in this study are fairly small (between 500 and 856 samples per domain). Efforts to release larger and more diverse data sets through multiple channels (such as challenges) in such domains as Alzheimer's  or aphasia detection, and depression detection~\cite{valstar2016avec, macwhinney2007talkbank, commonvoice} need to be reinforced.



\newpage
\bibliography{emnlp-ijcnlp-2019}
\bibliographystyle{acl_natbib}





\clearpage
\appendix

\section{List of Linguistic Features}
\label{app:feature_list}

\begin{table}[!htbp]
\begin{tabular}{l|l}
\textbf{Lexical Feature} & \textbf{Description} \\
\hline 
\hline 
distinct\_tokens & Number of distinct tokens \\
distinct\_tokens\_ratio & Number of distinct tokens occuring once \\
bigrams & Number of distinct bigrams \\
distinct\_bigrams & Number of distinct bigrams occuring once \\
distinct\_bigrams\_ratio & Ratio of distinct bigrams occuring once \\
trigrams & Number of distinct trigrams \\
distinct\_trigrams & Number of distinct trigrams occuring once \\
distinct\_trigrams\_ratio & Ratio of distinct trigrams occuring once \\
entropy\_1gram & Unigram entropy \\
entropy\_2gram & Bigram entropy \\
entropy\_3gram & Trigram entropy \\
cond\_entropy\_2gram & Conditional bigram entropy \\
cond\_entropy\_3gram & Conditional trigram entropy \\
wordtypes & Number of word types \\
swordtypes & Number of sophisticated word types \\
lextypes & Number of lexical types \\
slextypes & Number of sophisticated lexical word types \\
wordtokens & Number of word tokens \\
swordtokens & Number of sophisticated word tokens \\
lextokens & Number of lexical tokens \\
slextokens & Number of sophisticated lexical tokens \\
ld & Lexical density \\
ls1 & Lexical sophistication I \\
ls2 & Lexical sophistication II \\
vs1 & Verb sophistication I \\
vs2 & Verb sophistication II \\
cvs1 & Corrected VS1 \\
ndw & Number of different words \\
ndwz & NDW (first 50 words) \\
ndwerz & NDW (expected random 50) \\
ndwesz & NDW (expected sequence 50) \\
ttr & Type / token ratio \\
msttr & Mean segmental ttr (50) \\
cttr & Corrected ttr \\
rttr & Root ttr \\
logttr & Bilogarithmic ttr \\
uber & Uber coefficient \\
\hline
\end{tabular}
\end{table}

\clearpage

\begin{table}[h!]
\begin{tabular}{l|l}
\textbf{Syntactic Feature} & \textbf{Description} \\
\hline 
\hline 
S & Number of sentences \\
VP & Number of verb phrases \\
C & Number of clauses \\
T & Number of T-units\footnotemark \\
DC & Number of dependent clauses \\
CT & Number of complex T-units \\
CP & Number of coordinate phrases \\
CN & Number of complex nominals \\
MLS & Mean length of sentence \\
MLT & Mean length of T-units \\
MLC & Mean length of clause \\
C/S & Clauses per sentence \\
VP/T & Verb phrases per T-unit \\
C/T & Clauses per T-unit \\
DC/C & Dependent clauses per clause \\
DC/T & Dependent clauses per T-unit \\
T/S & T-units per sentence \\
CT/T & Complex T-units per T-unit \\
CP/T & Coordinate phrases per T-unit \\
CP/C & Coordinate phrases per clause \\
CN/T & Complex nominals per T-units \\
CN/C & Complex nominals per clause \\
\hline
\end{tabular}
\end{table}

\footnotetext{Here, T-unit is defined as the shortest grammatically allowable sentences into which writing can be split or minimally terminable unit. Often, but not always, a T-unit is a sentence.}

\end{document}